\documentclass[letterpaper]{article} % DO NOT CHANGE THIS
\usepackage{aaai23}  % DO NOT CHANGE THIS
\usepackage{times}  % DO NOT CHANGE THIS
\usepackage{helvet}  % DO NOT CHANGE THIS
\usepackage{courier}  % DO NOT CHANGE THIS
\usepackage[hyphens]{url}  % DO NOT CHANGE THIS
\usepackage{graphicx} % DO NOT CHANGE THIS
\usepackage{natbib}  % DO NOT CHANGE THIS AND DO NOT ADD ANY OPTIONS TO IT
\usepackage{caption} % DO NOT CHANGE THIS AND DO NOT ADD ANY OPTIONS TO IT
\usepackage{algorithm}
\usepackage{algorithmic}

\usepackage{newfloat}
\usepackage{listings}
\DeclareCaptionStyle{ruled}{labelfont=normalfont,labelsep=colon,strut=off} % DO NOT CHANGE THIS
\floatstyle{ruled}
\newfloat{listing}{tb}{lst}{}
\floatname{listing}{Listing}
\usepackage{multirow}
\usepackage{subcaption}
\usepackage{bm}
\usepackage{graphicx}
\usepackage{tabularx}
\usepackage{amsmath}
\usepackage{amsmath, amsthm, amssymb}
\usepackage{makecell}
\usepackage{rotating}
\usepackage{siunitx}
\usepackage{booktabs}

\usepackage[symbol]{footmisc}

\title{Prompt-Augmented Linear Probing: \\ Scaling beyond the Limit of Few-Shot In-Context Learners}

\author{Hyunsoo Cho\textsuperscript{\rm 1}, Hyuhng Joon Kim\textsuperscript{\rm 1}, Junyeob Kim\textsuperscript{\rm 1}, Sang-Woo Lee\textsuperscript{\rm 2, \rm3},\\ Sang-goo Lee\textsuperscript{\rm 1}, Kang Min Yoo\textsuperscript{\rm 1, \rm2,}$^*$, Taeuk Kim\textsuperscript{\rm 4,}\thanks{Corresponding authors.}}
\affiliations{
    \textsuperscript{\rm 1} Seoul National University \\
    \textsuperscript{\rm 2} NAVER Cloud \\
    \textsuperscript{\rm 3} KAIST\\
    \textsuperscript{\rm 4} Hanyang University\\
    \{johyunsoo, heyjoonkim, juny116, sglee\}@europa.snu.ac.kr \\
    \{kangmin.yoo, sang.woo.lee\}@navercorp.com, 
    kimtaeuk@hanyang.ac.kr \\
}
\begin{document}
    \maketitle
    \begin{abstract}
  
    Through in-context learning (ICL), large-scale language models are effective few-shot learners without additional model fine-tuning. 
    However, the ICL performance does not scale well with the number of available training samples as it is limited by the inherent input length constraint of the underlying language model. 
    Meanwhile, many studies have revealed that language models are also powerful feature extractors, allowing them to be utilized in a \textit{black-box} manner and enabling the linear probing paradigm, where lightweight discriminators are trained on top of the pre-extracted input representations.
    This paper proposes prompt-augmented linear probing (PALP), a hybrid of linear probing and ICL, which leverages the best of both worlds.
    PALP inherits the scalability of linear probing and the capability of enforcing language models to derive more meaningful representations via tailoring input into a more conceivable form.
    Throughout in-depth investigations on various datasets, we verified that PALP significantly enhances the input representations closing the gap between ICL in the data-hungry scenario and fine-tuning in the data-abundant scenario with little training overhead, potentially making PALP a strong alternative in a \textit{black-box} scenario. 
    % We release our code at \url{https://github.com/HyunsooCho77/PALP}.
\end{abstract}
    \section{Introduction}
    Since the emergence of Transformer-based \cite{vaswani2017attention} language models, we have witnessed notable improvements in the natural language processing literature, even attaining human-level performance on several benchmarks.
    In addition, as it becomes evident that scaling laws work for such models \cite{kaplan2020scaling}, there has been a significant amount of investment in the field to enhance them in terms of the number of their parameters---from millions to billions---and the volume of the data they consume during training \cite{brown2020language, chowdhery2022palm, fedus2021switch,hoffmann2022training}.
    As a result, some cutting-edge language models become possible to obtain intriguing extra functionalities, such as the ability to capture world knowledge \cite{petroni2019language}, generate codes \cite{poesia2022synchromesh}, or solve mathematical problems \cite{henighan2020scaling}, in addition to being proficient in recognizing linguistic patterns.
    These anecdotes, which demonstrate the general power of large language models, naturally raise researchers' expectations that language models can act as a universal, off-the-shelf solution for a range of downstream tasks while minimizing the cost required for adapting them to a specific job at the same time.

\begin{figure}[t]
    \begin{center}
        \includegraphics[width=1\columnwidth]{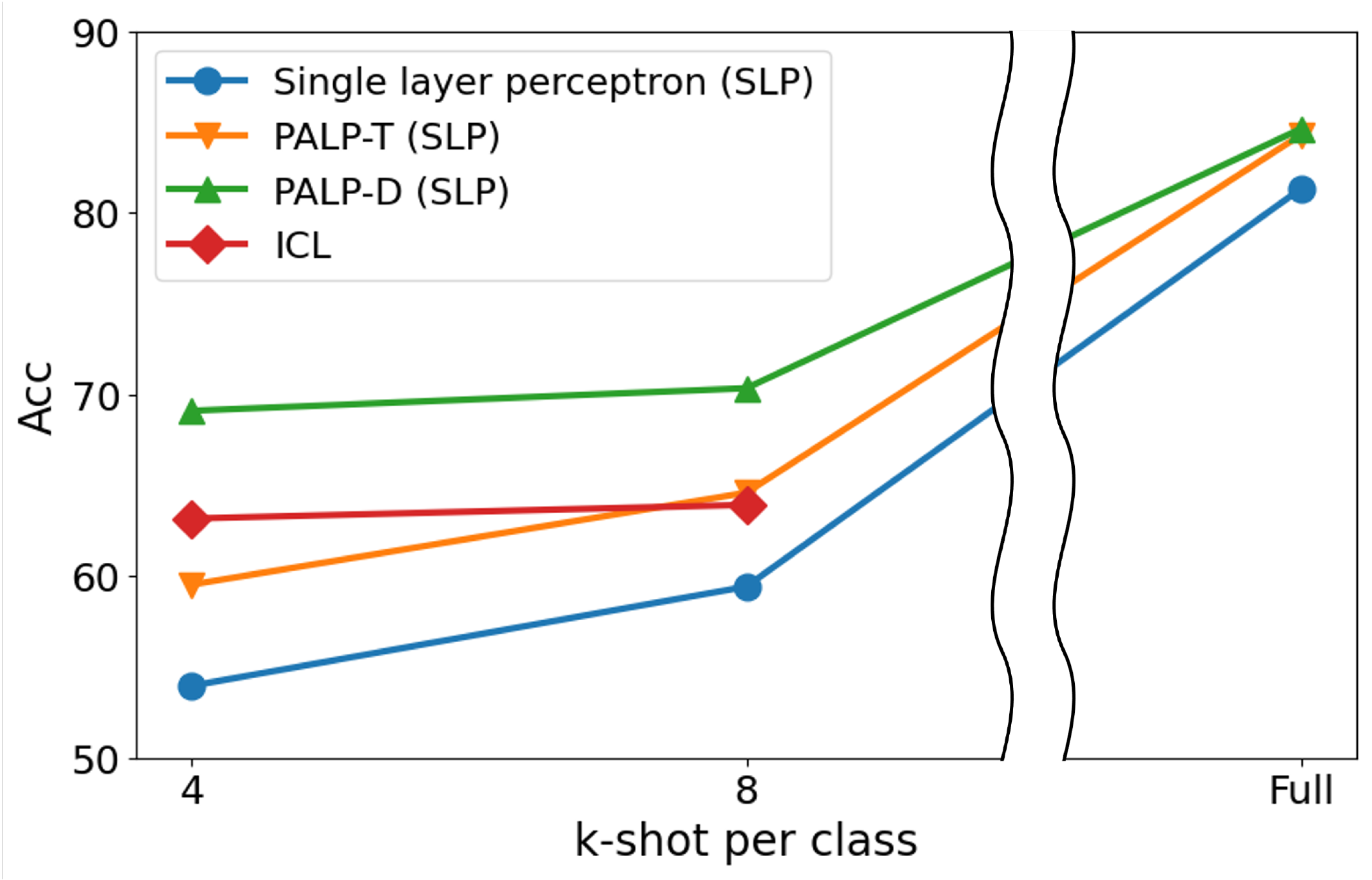}
          \caption{Average accuracy (12 classification tasks) of various \textit{black-box} transferring methods trained on GPT-J.
          ICL can not leverage the full train dataset due to the length limit, and their performance saturates quickly.
          Our method (PALP) is scalable with available training samples and minimizes the performance gap between ICL in a few-shot setting.
          The performance of the respective task is summarized in Table \ref{tab:few-shot_4_8}.
          }
          \label{fig:front_image}
    \end{center}
\end{figure}

    However, there is no free lunch; the effectiveness and generalizability of large language models achieved by scaling come at the cost of physically serving such gigantic neural architectures.
    Thus, the institutions that distribute large models such as GPT-3 \cite{brown2020language}  usually pursue the strategy of providing commercial APIs which only allow limited access to the models.
    In other words, it is often the case that users cannot receive information about the inner workings of the models, such as gradients concerning the models' parameters which are crucial for fine-tuning the models for a particular purpose.
    Therefore, there has been a growing interest in adapting language models in this restricted setting, dubbed as \textit{black-box tuning} \cite{sun2022black, diao2022black}.

    In this scenario, especially focusing on classification, the two mainstream paradigms are linear models and in-context learning (ICL).
    The former derives the final representation from a language model and trains a simple classifier, e.g., Support Vector Machine (SVM), Single-Layer Perceptron (SLP), and $k$ Nearest Neighbors ($k$-NN), on top of the extracted features.
    On the other hand, ICL is a training-free mechanism that makes the most use of the nature of language models.
    Specifically, ICL promotes a language model to generate the desired output by guiding the model with a few examples of the target task (i.e., demonstrations) plus a set of templates tailored for the task, where many works have suggested that exploiting such a mechanism improved various NLP tasks \cite{shwartz2020unsupervised, ben2021pada}.
    
    Yet, it is worth noting that according to the number of data instances available for training, the two aforementioned approaches have clear strengths and shortcomings.
    Namely, ICL shows a remarkable generalization ability in a few-shot setting. 
    However, the ICL performance does not scale well with the number of available training samples (See Figure \ref{fig:front_image} for details).
    This is because ICL relies on the innate capability of language models, i.e., being able to predict the next word given context, and a verbalizer, which is a function that maps the probabilities of some pre-defined tokens to the probability of each label rather than constructing a task-specific decision maker.
    By doing so, ICL is equipped with the invaluable benefit of being free from explicit fine-tuning, but it is also a double-edged sword that induces a scalability issue.
    In particular, an explicit upper bound exists on the performance of ICL, which is determined by the maximum length of the utilized language model.
    
    In this paper, we show that the combination of linear classifiers and the techniques introduced for ICL can cover the weaknesses of each method.
    We train diverse linear classifiers whose representations are extracted from input pre-processing strategies invented for facilitating ICL.
    Specifically, we augment training data instances with the templates or prepend additional demonstrations in front of the input of interest for better contextualization.
    
    We validate our method with various datasets, demonstrating that it is consistently superior to baselines in both the low-data and full-data settings.
    From empirical experiments, we observe that exploiting templates that provide hints about the target task or concatenating demonstrations can significantly enhance the extracted representations from PLM, improving the classifiers' performance in various scenarios and reducing the gap between ICL and fine-tuning.
    Intriguingly, we also discover that importing the techniques directly from ICL without care may cause the inheritance of the disadvantages of ICL, such as a substantial performance variance depending on the appended demonstrations or high sensitivity to the format of templates.

\begin{figure*}
    \begin{center}
        \includegraphics[width=0.99\textwidth]{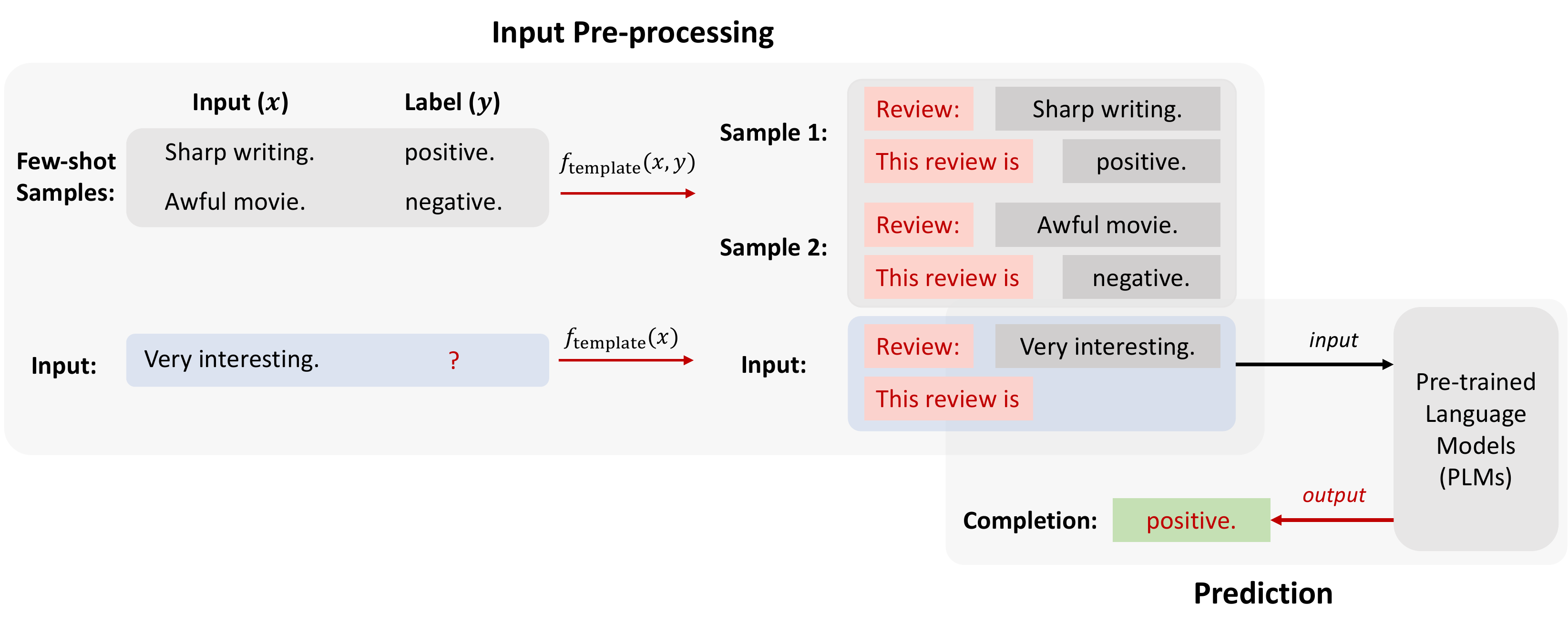}
          \caption{Overall illustration of ICL. ICL utilizes a template function to convert available few-shot samples into demonstrations $f_{\text{templtate}}(x, y)$, and appends the generated demonstrations to the front of the \textit{templified} current input $f_{\text{templtate}}(x)$. Then, ICL uses LM's prediction ability from the pre-training step to infer the correct answer.}
          \label{fig:icl_illustration}
    \end{center}
\end{figure*}

\section{Preliminary}

\subsection{Problem Formulation}
    In this work, we consider classifying input sequences based on \textit{black-box tuning} \cite{sun2022black, diao2022black}, where the parameters of pre-trained language models (PLMs) are inaccessible.
    That is, PLMs only serve as an encoder function $e$ that delivers $n$-dimensional latent features $\bm{h}$ from input $\bm{x}$.
    Formally, for input $\bm{x}\in \mathcal{X}$, let the $n$-dimensional continuous latent features extracted from PLMs be $\bm{h}=e(\bm{x})$, where  $\bm{h}\in\mathcal{H}$.
    In addition, let $\mathcal{Y}=\{0,1,\cdots,|C|\}$ be a label space, where $|C|$ is the cardinality of the space.
    Then, a classifier $p(y|\bm{h};\bm{\theta}):\mathcal{H}\rightarrow\mathcal{Y}$ maps $\bm{h}$ to a class label $y\in\mathcal{Y}$, estimating the probability of input $\bm{x}$ belonging to a certain label.

\subsection{Linear Classifiers}
    Linear models, such as single layer perceptron (SLP), Support Vector Machine (SVM), or logistic regression (LR), are trained by solving optimization problems concerning their parameters.
    First-order methods such as gradient descent shown below are a general choice for parameter estimation:
    \begin{equation}
        \bm{\theta}_{t+1}=\bm{\theta}_{t} - \eta \nabla\mathcal{L}(\bm{\theta}(t),\mathcal{H}_{batch}),
        \label{eq:sgd}
    \end{equation}
    where $\mathcal{L}$, $\eta$, $\mathcal{H}_{batch}$ refer to a loss function (e.g., the cross-entropy loss, hinge loss in SVM, and MSE loss in regression), learning rate, and a mini-batch sampled from the training dataset.
    In this paper, we evaluate 5 different linear-probing classification methods: $k$-NN, LR, SVM, gaussian discriminative analysis (GDA), and SLP.
    The details of the mentioned approaches are presented in the Appendix.

\subsection{In-Context Learning}
    In-context learning (ICL) is a brand new, training-free paradigm that attempts to make the most use of the nature of language models to conduct a target task.
    ICL promotes a language model to generate the desired output by guiding the model with a few examples of the target task (i.e., demonstrations) plus a set of templates tailored for the task.
    
    In detail, ICL consists of two steps \cite{liu2021pre}: 
    First, the \textbf{input pre-processing} step combines the input of interest $\bm{x}$ with $k$-shot samples (i.e., demonstrations) $(\bm{x}_{i}$,$y_{i})_{i=1}^{k}$ from the training set.
    Then, a template function $f_{\text{template}}(\cdot)$ attach pre-defined descriptions to the input $f_{\text{template}}(\bm{x})$ or additionally attach the corresponding natural language label to the \textit{templified} input $f_{\text{template}}(\bm{x},y$), commonly referred to as a demonstration. 
    (See Figure \ref{fig:icl_illustration} for a graphical explanation.)
    
    For instance, the function attaches a prefix or postfix to the original $\bm{x}$, or it transforms $y_{i}$ into the form of natural language rather than numeric numbers.
    In consequence, the final input for ICL, denoted as $\bm{\hat{x}}$, becomes a concatenation of all the pre-processed $\bm{x}$ and $(\bm{x}_{i}$,$y_{i})_{i=1}^{k}$:
    
    \begin{equation}
        \bm{\hat{x}} = D_{1}\oplus D_{2}\oplus\cdots \oplus D_{k}\oplus f_{\text{template}}(\bm{x}),
    \end{equation}
    where $D_{i}=f_{\text{template}}(\bm{x}_i,y_{i})$ and $\oplus$ refers to the concatenation operation.

    Second, in the \textbf{prediction} phase, ICL leverages PLMs to compute the feature $\bm{\hat{h}}= e(\bm{\hat{x}})$, followed by a \textit{verbalizer} $\mathcal{V}: \mathcal{H} \rightarrow \mathcal{Y}$ that is a reformulation of the language model head for task-specific adaptation.
    It is often assumed that the verbalizer only considers single tokens as its output candidates, which correspond to each item in the label space $\mathcal{Y}$.

\section{Prompt-Augmented Linear Probing (PALP)}

    \subsection{Motivation}
        
        The primary intuition behind our method borrows from the in-context learning ability exhibited by language models. 
        Specifically, in-context learners benefit from more elaborate and longer prompts \cite{reynolds2021prompt}, allowing them to carry out deeper reasoning through longer input sequences and corresponding hidden layers.
        
        We posit that providing appropriate guidance to the language model via prompts (input pre-processing step in ICL) benefits not only the usual causal language modeling (i.e., predicting the next token) ability but also enhances the quality of the representation for the input text.
        The primary goal of our method is to extract a more distinctive representation from PLMs via crafting a raw dataset into a more understandable form and training linear probers on the top of the extracted representations.
        Specifically, we transform the dataset in two ways:
        
        \begin{enumerate}
            \item We utilize a simple template that gives a brief description of input and the objective of the task.
            \item On top of \textit{templified} dataset, we concatenate a single class-wise demonstration to an inferring input to give important cues, i.e., input-label mapping, label space, or distribution of the input text \cite{min2022rethinking} regarding the target task. 
        \end{enumerate}

        In the following subsection, we explain the dataset reconstruction strategies for each method.

    \subsection{PALP-Template (PALP-T)}
        Applying a template to the input is the most straightforward and intuitive way to enforce PLM to follow user requirements.
        Accordingly, we attempt to extract more task-specific features by attaching a fixed prefix, infix (for sentence pair tasks), or postfix, which describes a raw input into a more understandable form.

        For instance, we transform sentiment analysis instance `\textit{very interesting.}' into `\textbf{Review:} \textit{very interesting.} \textbf{Sentiment:}' to provide the language model additional indications that the input is the form of review, and the user expects sentiment as an answer.
        Formally, for the training dataset $\mathcal{D}_{\text{train}}=\{(\bm{x}_i,y_i)| i\in m\}$, we convert this into 
        $\mathcal{D}_{\text{train}}^{\text{template}} = \{(f_{\text{template}}(\bm{x}_i),y_i)| i\in m\}$, 
        where $f_{\text{template}}(\cdot)$ is a template function.
        Then we train linear classifiers with transformed dataset $\mathcal{D}_{\text{train}}^{\text{template}}$ in the exact same way as in Eq. \ref{eq:sgd}.
        
        In the inference time, we also have to apply the same template function to inferring input $f_{\text{template}}(\bm{x}_{\text{test}})$ in order to match the format.

\begin{figure*}
    \begin{center}
        \includegraphics[width=0.9\textwidth]{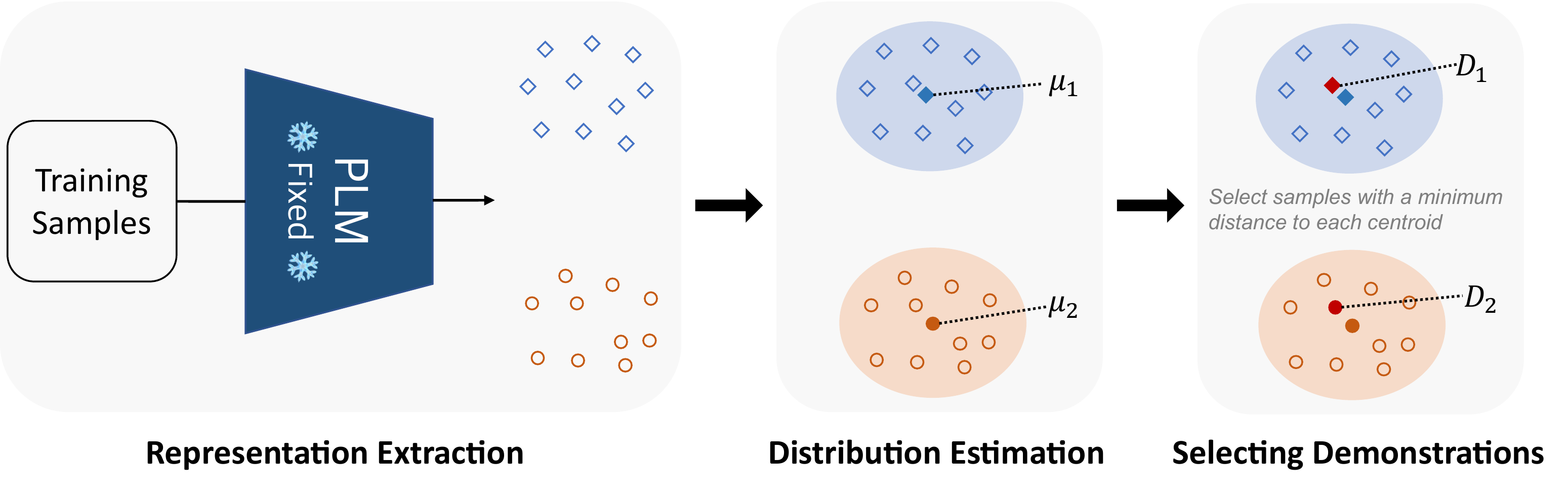}
          \caption{Illustration of selecting demonstrations in PALP-D in the binary classification task. We estimate the normal distribution for each class from available training samples and select the closest sample, respectively.}
          \label{fig:demonstration_selection}
    \end{center}
\end{figure*}

    \subsection{PALP-Demonstration (PALP-D)}
        On top of the previous \textit{templified} dataset (PALP-T), PALP-D additionally concatenates some demonstrations to the front to maximize the capability of PLM to learn \textit{in-context}.
        However, unlike ICL, which attaches all available samples, our method extracts a single representative demonstration per class (i.e., total $|C|$ demonstrations) and leverages them as demonstrations.
        Formally, let there exist $k$ accessible training samples per each class label:
        \begin{equation}
            \mathcal{D}_{\text{train}} = \{(\bm{x_j^i},y_j^i)|j\in k,i\in|C|\},
        \end{equation}
                \noindent where $|C|$ is the cardinality.
        Then, we choose one instance from each class and concatenate them into one prefix $\tau$:
        \begin{align}
                \bm{\hat{x}} &= \tau \oplus f_{\text{template}}(\bm{x}), \text{ where}\\
                 \tau &= D_{1}\oplus, \cdots, \oplus D_{|C|}.
        \end{align}

        \noindent In this way, we can avoid atrociously lengthy input, minimizing the computational cost and enhancing the method's scalability.
        Specifically, while the length of the input in ICL is normally proportional to the total available number of samples, PALP-D is proportional to the number of labels of the task.
        
        Given that PALP-D does not inject complete available training samples, the core is selecting a single meaningful demonstration that can distill PLM sufficient knowledge to comprehend essential signs of the task, such as input-label mapping, label space, or distribution of the input text.
        We hypothesize that the inputs closest to the centroid of each class label can capture the most representative information for respective classes and extract them as a set of demonstrations $P=\{D_{1},D_{2},\cdots,D_{|C|}\}$.

        In order to do so, we first estimate the normal distribution for each class label $\mathcal{N} (\mu_{i \in |C|},\Sigma_{i \in |C|})$ from the available training inputs and measure the distance between entire training samples and the estimated centroid through Mahalanobis distance:
        \begin{equation}
            \label{eq:mahalanobis-score}
            dist_{\text{Mahal}}(
                \bm{x},
                \mu_i;
                \Sigma_i
            ) =
            (\bm{x} - \mu_i)^\top
            \Sigma^{-1}_i
            (\bm{x} - \mu_i)
            .
        \end{equation}
        Finally, we select samples closest to each class centroid and utilize them as demonstrations $D_i = \min_{j\in k}(x_j^i, u_i;\Sigma_i)$.
        (Figure \ref{fig:demonstration_selection} illustrates the selection of demonstration from available training samples.)
        
        In a few-shot scenario, we randomly permuted the order of available demonstrations $P$ to construct multiple prefixes $\hat{\tau}$ in the training phase, where $\hat{\tau} \sim \sigma(P)$ and $\sigma(\cdot)$ refers to a random permutation operator. 
        By doing so, we can generate $|C|!$ different prefixes that can give the effect of data augmentation and alleviate the data-scarcity problem.
        And in the inference phase or data-abundant setting, we attach a unified prefix $\tau$ (without random permutation) to the front of the test input to match the format with the training samples.

\begin{table}[t!]
    \centering
    \resizebox{1 \columnwidth}{!}{%
    \begin{tabular}{c|l|c|c|c}
        \toprule
            Task type & Dataset & \# Class & \# Train & \# Test \\
        \midrule
            \multirow{10}{*}{Single Sentence} &SST2 & 2& 67,349 & 872 \\
            & Rotten\_tomatoes &2& 8,530 & 1,066 \\
            & Offensive &2& 19,916 & 860 \\
            &CoLA &2& 8,551 & 1,043 \\
            &Stance\_atheism &3& 461 & 220 \\
            &Emotion &4& 3,257 & 1,421 \\
            &AG\_news &4& 120,000 & 7,600 \\
            &TREC &6& 5,452 & 500 \\
            &Banking77 & 77 & 10,003 & 3,080 \\
            &CLINC & 150 & 15,000 & 4,500 \\
        \midrule
            \multirow{5}{*}{Sentence pair}&MNLI &3& 392,702 & 9,832 \\%(val-matched)
            &MRPC &2& 3,668 & 408 \\
            &RTE &2& 2,490 & 277 \\
            &BoolQ &2& 9,427 & 3,270 \\
            &CB &3& 250 & 56 \\
            % QNLI &$\pm$2& 104,743 & 5,463 \\ 
        \bottomrule
    \end{tabular}
    }
    \caption{Statistics of 15 different datasets used in our experiments.}
    \label{tab:dataset_stat}
\end{table}

\settowidth\rotheadsize{Minimal}
\begin{table*}[t!]
\centering
% \resizebox{2.1 \columnwidth}{!}{
    \small
    \begin{tabular}{c|c|ccccccccccc|c} 
    \toprule
    \multicolumn{14}{c}{GPT-J 4-shot per class}                                            \\ 
    \midrule
    \multicolumn{2}{c|}{Method}   & AG & SST2 & Rotten & Stance & Emotion & TREC & CoLA & Offensive & MNLI & RTE & MRPC & AVG \\ 
    \midrule
    \multirow{5}{*}{B} & k-NN & 54.66 & 51.33 & 58.67  & 54.73  & 33.92  & 48.84  & 42.36  & 49.37 & 35.09 & 48.74 & 64.66 & 49.31 \\
                      & LR & 66.48 & 50.25 & 65.08 & 58.64 & 36.89 & 64.48 & 50.68 & 51.72 & 36.46 & 49.39 & 59.02 & 53.55 \\
                      & SVM & 66.20 & 50.41  & 65.91 & 59.27 & 36.59 & 65.92  & 48.90 & 48.72 & 36.02 & 49.31 & 61.96 & 53.56\\
                      &SLP & 67.88 & 50.23 & 65.68 & 59.45 & 37.13 & 66.80 & 49.13 & 49.95  & 36.36  & 49.24  & 61.76 & 53.96 \\
                      & GDA & 66.32  & 50.41  & 65.93  & 58.64  & 36.88  & 66.44  & 48.88  & 48.72  & 36.03  & 49.24  & 61.96 & 53.59 \\
    \midrule
    \multirow{5}{*}{T} & k-NN & 61.88   & 58.30  & 65.10   & 35.27  & 45.18  & 51.04  & 53.02  & 59.98  & 36.40  & 53.14   & 61.47 & 52.80 \\ 
                      & LR & 71.84   & 62.00  & 71.73   & 58.09  & 50.56   & 65.56  & 49.86  & 59.00  & 38.73   & 50.54   & 60.44 & 58.03  \\
                      & SVM & 72.08   & 63.67  & 71.33   & 56.27   & 50.27   & 66.36  & 53.08  & 62.02 & 37.93   & 49.60   & 61.18 & 58.53 \\
                      &SLP & 73.79   & 63.58  & 72.18   & 57.45   & 52.71   & \underline{68.44} & \underline{58.81} & 63.07  & 38.25   & 49.96   & 61.57  & 59.53 \\ 
                      & GDA & 72.82   & 64.20  & 71.33   & 56.27   & 52.86   & 66.00  & 53.08  & 62.02  & 38.13  & 49.60   & 61.18 & 58.86  \\ 
    \midrule
    \multirow{5}{*}{D} & k-NN & 75.99  & 70.16  & 71.01  & 49.45  & 56.95  & 46.80  & 55.67  & 54.56  & 38.85 & 51.09  & 63.86 & 57.67 \\
                        & LR & 77.92  & 70.62  & 80.58 & \underline{61.27} & 68.97  & 65.96  & 53.83  & 67.63  & \textbf{40.30} & 49.00  & 62.27 & 63.49 \\
                        & SVM & 77.96  & 75.46  & 79.42  & 55.82  & 68.91  & 66.24 & 55.30  & 63.75  & \underline{40.12} & 51.54  & 64.76 & \underline{63.57} \\
                        &SLP & \underline{80.69} & \underline{77.41} & \underline{81.01}  & \textbf{71.36} & \underline{70.38} & \textbf{71.92} & \textbf{69.13} & \underline{72.35} & 39.41  & \underline{54.66} & \textbf{71.73} & \textbf{69.10}\\
                        & GDA & 76.08  & 70.16 & 68.05  & 55.91  & 65.66  & 63.76  & 54.42  & 69.18  & 39.32  & 50.53  & 64.15 & 61.57 \\
    \midrule
    \multicolumn{2}{c|}{ICL}      & \textbf{81.74} & \textbf{91.77 } & \textbf{90.45} & 23.09  & \textbf{72.76} & 64.00  & 37.89  & \textbf{73.19} & 36.04  & \textbf{55.60} & \underline{68.38} & 63.17 \\
    \midrule
    \multicolumn{14}{c}{GPT-J 8-shot per class}                                                                                                                                                                                                                                                                                  \\ 
    % \hline
    % \multicolumn{2}{c|}{Method}   & TREC                 & CoLA                 & SST-2                & Offensive            & Stance               & Emotion              & AGNews               & RottenTomatoes       & MNLI                 & RTE                  & MRPC                   & AVG                  \\ 
    \midrule
    \multirow{5}{*}{B} & k-NN & 63.05  & 50.55  & 65.42  & 58.09  & 36.05  & 60.16  & 52.41  & 52.26 & 35.69  & 51.70   & 59.17 & 53.14 \\
                        & LR & 74.56  & 57.50  & 74.15  & 65.73  & 41.90  & 79.08  & 52.79  & 58.47  & 37.43   & 51.34  & 58.97 & 59.27\\
                        & SVM & 73.26  & 57.16  & 75.87  & 64.27   & 42.66  & 80.12  & 53.98  & 57.79  & 37.43   & 52.06  & 59.56  & 59.47\\
                        &SLP & 74.72  & 57.91  & 75.25  & 63.82  & 42.87  & 80.92  & 54.34  & 55.93  & 37.37  & 51.05  & 59.46  & 59.42\\ 
                        & GDA & 74.12  & 57.16  & 75.78  & 63.00  & 44.31  & \textbf{81.84} & 53.98  & 57.79  & 37.60  & 52.06  & 59.56 & 59.75 \\
    \midrule
    \multirow{5}{*}{T} & k-NN & 69.65  & 61.93  & 65.46  & 56.45  & 49.87  & 62.44  & 45.29  & 56.35  & 38.13  & 50.90  & 56.42  & 55.72 \\
                      & LR & 78.52  & 69.33  & 77.04  & 66.00  & 59.93  & 79.04  & 51.51  & 59.30  & 41.24  & 52.56  & 63.48  & 63.45 \\
                      & SVM & 78.39  & 75.44  & 77.97  & 63.36  & 60.53  & 81.12  & 51.98  & 65.07  & 40.45   & 53.07  & 61.13  & 64.41 \\
                      &SLP & 79.61  & 72.80  & 78.03  & 66.09 & 62.36  & 80.12  & 52.10  & 64.37  & 40.87  & 52.49  & 61.67  & 64.59 \\
                      & GDA & 79.03  & 75.37  & 77.90  & 62.55  & 62.15  & \underline{81.68} & 51.98  & 65.05  & 40.90  & 53.07  & 61.18  & 64.62 \\
    \midrule
    \multirow{5}{*}{D} & k-NN & 79.21  & 67.98  & 78.82  & 49.63  & 58.31  & 58.36  & 50.43  & 63.26  & 38.24  & 51.37  & 56.94  & 59.32 \\
                        & LR & \underline{84.12}  & 73.78  & 85.33  & \underline{66.55}  & 66.66  & 69.04 & \underline{57.20}  & 69.21 & 42.25  & 53.43  & 64.30   & 66.53 \\
                        & SVM & 83.96 & 77.10 & 85.27  & 64.36  & 68.32  & 71.20  & 55.55  & \underline{71.65}  & \textbf{43.00} & \underline{53.60} & 60.82   &  \underline{66.80}  \\
                        &SLP & \textbf{85.27} & \underline{78.37} & \underline{86.75} & \textbf{69.27} & \underline{69.97} & 75.76  & \textbf{69.63} & 71.23  & \underline{42.30} & 53.26  & \textbf{71.94}  & \textbf{70.34} \\
                        & GDA & 83.05  & 72.55  & 83.83  & 61.18  & 64.32  & 68.08  & 55.02 & 71.05  & 42.10  & 51.42  & 61.92  &  64.96 \\ 
    \midrule
    \multicolumn{2}{c|}{ICL}      & 83.26  & \textbf{91.72} & \textbf{89.72} & 27.27  & \textbf{73.12} & 71.60  & 34.28  & \textbf{73.02} & 36.62  & \textbf{54.08} & \underline{68.38 }  & 63.92 \\
    \bottomrule
    \end{tabular}
    
% }
\caption{Experimental results of 11 different datasets on GPT-J in 4-shot / 8-shot per class settings. B, T, and D refers to a baseline, PALP-Template, and PALP-Demonstration respectively. For each dataset, the best method is in bold and the second best method and is underlined. }

\label{tab:few-shot_4_8}
\end{table*}

\section{Experiments}
\label{sec:experiments}

\subsection{Experimental Setup}
    \noindent\textbf{Backbone. }In the following experiments, we adopt GPT-J \cite{gpt-j} as the main backbone of our experiments, with additional experiments using GPT-2 \cite{radford2019language}.
    
    \noindent\textbf{Datasets. } To investigate the performance of each method in many different scenarios, we select 15 datasets, as stipulated in Table \ref{tab:dataset_stat}.
    The selected dataset covers single sentence task to sentence pair tasks, binary to 150 classes (various numbers of class labels), in diverse domains.
    The detailed list of each dataset and references are covered in Appendix.

\settowidth\rotheadsize{Minimal}
\begin{table*}[t]
\centering
\resizebox{2.1 \columnwidth}{!}{
    \begin{tabular}{c|c|ccccccccccc|c} 
    \toprule
    \multicolumn{14}{c}{GPT-J Full dataset} \\ 
    \midrule
    \multicolumn{2}{c|}{Method} & AG & SST2 & Rotten & Stance & Emotion & TREC & CoLA & Offensive & MNLI & RTE & MRPC & AVG\\ 
    \midrule
    \multirow{5}{*}{B} & k-NN & 88.87 & 76.15 & 84.99 & 68.64 & 56.72 & 91.80 & 66.73 & 74.42 & 43.63 & 51.26 & 67.89 & 70.10 \\
                     & LR  & 90.96 & 91.97 & 86.96 & 72.32 & 72.20 & \textbf{97.60} & 77.37 & 73.60 & 71.07 & 68.59 & 75.22 & 79.81\\
                     & SVM & 90.34 & 90.48 & 88.18 & 72.27 & 71.26 & 96.80 & 76.13 & 72.91 & 67.32 & 69.68 & 75.98 & 79.21\\
                     &SLP  & 90.43 & 90.88 & 89.51 & 75.00 & 75.40 & 97.00 & 77.87 & 80.69 & 70.70 & 70.71 & 76.44 & 81.33\\
                     & GDA & 92.18 & 92.40 & 89.49 & 72.27 & 72.56 & 97.00 & 78.52 & 74.07 & 71.07 & 68.59 & 74.27 & 80.22\\ 
    \midrule
    \multirow{5}{*}{T} & k-NN & 89.87 & 87.61 & 85.83 & 66.82 & 72.55 & 91.60 & 66.16 & 73.84 & 51.70 & 57.40 & 73.04 & 74.22\\
                       & LR   & 92.58 & 92.66 & 88.84 & 77.73 & 79.80 & 96.60 & 78.81 & 80.00 & 76.28 & 73.65 & 80.39 & 83.39\\
                       & SVM  & 90.58 & 89.45 & 85.83 & 75.91 & 79.52 & 97.40 & 75.10 & 73.84 & 72.58 & 71.48 & 80.39 & 81.10\\
                       &SLP   & 92.49 & 93.35 & 89.87 & 78.36 & 81.30 & \textbf{97.60} & 79.19 & 82.67 & 77.41 & 72.20 & 82.84 & 84.30\\
                       & GDA  & 92.36 & \underline{94.27} & 90.81 & 75.46 & 81.07 & 97.00 & 78.91 & 82.21 & 75.91 & 71.84 & 79.17 & 83.55\\ 
    \midrule
    \multirow{5}{*}{D} & k-NN & 90.69 & 91.17 & 89.96 & 75.00 & 73.89 & 89.40 & 69.70 & 77.09 & 51.54 & 54.51 & 72.55 & 75.95 \\
                       & LR   & 92.47 & 92.73 & 90.54 & 77.73 & 79.73 & 95.40 & 80.25 & 82.79 & 71.94 & 76.17 & 76.96 & 83.34\\
                       & SVM  & 90.78 & 91.97 & 88.37 & 77.27 & 76.92 & 95.60 & 80.06 & 75.47 & 51.38 & 76.53 & 76.23 & 80.05\\
                       &SLP   & 92.58 & 93.37 & \underline{91.33} & \textbf{83.51} & 82.31 & 94.00 & 77.31 & 82.23 & 76.28 & 77.62 & 80.39 & 84.63\\
                       & GDA  & 92.86 & 93.00 & 90.81 & 73.18 & 81.66 & 96.60 & 79.10 & 81.16 & 75.26 & 77.26 & 77.70 & 83.51\\ 
    \midrule
    \multicolumn{2}{c|}{Adapter}     & \textbf{95.50} & \textbf{95.53} & 90.26 & \underline{81.53} & \underline{83.54} & \underline{97.41} & \textbf{84.48} & \textbf{84.67} & \textbf{88.84} & \underline{82.80} & \textbf{88.48} & \textbf{88.46}\\
    \midrule
    \multicolumn{2}{c|}{Fine-tuning} & \underline{94.80} & 94.15 & \textbf{91.79} & 81.25 & \textbf{84.41} & 97.22 & \underline{82.34} & \underline{84.02} & \underline{87.47} & \textbf{83.33} & \underline{86.51} & \underline{87.94}\\
    \bottomrule
    \end{tabular}
}
\caption{Experimental results of 11 different datasets on GPT-J in full datases settings. B, T, and D refers to a baseline, PALP-Template, and PALP-Demonstration respectively. For each dataset, the best method is in bold and the second best method and is underlined.}
\label{tab:full-shot}
\end{table*}

    \noindent\textbf{Setting \& Reporting Methods. }Our experiments cover both a few-shot setting and full training data setting.
    As a reporting method, we select currently prevailing linear probing methodologies (i.e., SVM, SLP, LR, GDA, and $k$-NN) and their application of PALP.
    In a few-shot setting, we take a closer look at how the performance of each method changes when the training input is converted into ICL style and compare them with ICL, which is known to yield superior performance in the data-hungry setting.
    In the full-shot setting, we investigate the performance gap between our method and the \textit{white-box} training method (i.e., accessible to model parameters), such as Adapter \cite{houlsby2019parameter} or full fine-tuning, which can be considered upper bound.

    \noindent\textbf{Other details.}
    We optimized the hyper-parameters of each classification method on SST2 dataset with 4 Tesla V100 SXM2 32GB GPUs and universally utilized them in different settings. 
    (Detailed hyper-parameters and implementations are in the Appendix.)
    Additionally, we found a manual template for each task where ICL exhibited sound performance and utilized them universally in our methods. (All templates for each task are stipulated in the Appendix.)
    For stable evaluation, we report the average of 5 different seeds (13, 27, 250, 583, 915) as a model performance and report standard deviations for each task in the Appendix.

\subsection{Few-shot Results}
    In the few-shot setting, we experimented based on the number of accessible samples for each task class.
    For instance, a 4-shot setting in the Sentiment Analysis task with 2 classes (positive and negative) means that a total of 8 samples are accessible, which is analogous to a balanced 8-shot setting in ICL.
    Table \ref{tab:few-shot_4_8} summarizes the performance of ICL, baseline linear probing methods, and their application of PALP (T and D) in the 4,8-shot setting.
    Baseline refers to utilizing raw input without modification, which is a conventional supervised learning paradigm.
    Template (PALP-T) and demonstration (PALP-D) refer to template-based training samples and demonstration-based training samples from our method individually.
    We now refer to a T for PALP-T and D for PALP-D for short. (Additional results with 16-shot is in the Appendix)

    While ICL displays sound performance in various tasks, the baseline linear probing method exhibits poor performance compared to ICL. 
    In sentiment analysis tasks (SST2, rotten tomatoes), the performance of ICL is above 90\% only with 4-shot samples per class, whereas the baseline linear probing methodology almost makes arbitrary random decisions in the same environment (around 50\% $\sim$ 60\%).
    However, the performance of ICL  quickly saturates and scales poorly with the number of available training samples. 
    And even in some cases, ICL performs worse than arbitrary decisions without understanding the target task at all (e.g., stance, CoLA).
    Furthermore, if the input length exceeds a certain level, it is infeasible to utilize ICL in the usual way. (See the 16, 32-shot results in the Appendix.)
    On the other hand, linear probing methods are much more scalable with the number of available samples, revealing stable performance regardless of the dataset. 
    Moreover, their application of PALP boosts performance by a substantial margin minimizing the gap between ICL, especially in most single-sentence tasks.
    We can obtain around 5\% improvement in average from each ablation (appending template and demonstrations) in the 4-shot setting, and some linear probing methods outperform ICL.
    The most high-performance results were obtained from the SLP among other linear probing methodologies.

\begingroup
\setlength{\tabcolsep}{8pt} % Default value: 6pt

\begin{table}[t]
\centering
\resizebox{1 \columnwidth}{!}{%
    % \small
    \begin{tabular}{c|c|cccc}
    \toprule
        \multicolumn{6}{c}{GPT-J Full train dataset} \\
    \midrule
    \multicolumn{2}{c|}{Method} & CLINC & Banking & CB & BoolQ \\
    \midrule
        \multirow{5}{*}{B} & k-NN & 74.78 & 69.06 & 71.43 & 63.01 \\
         & LR & 92.91  & 89.44 & 80.36 & 62.70 \\
         & SVM & 91.20  & 89.55 & 80.36 & 63.77 \\
         &SLP & 91.52 & 89.43 & 80.35 & 62.70 \\
         & GDA & 93.78 & 89.54 & 80.36 & 63.00 \\
    \midrule
        \multirow{5}{*}{T} & k-NN & 90.42 & 86.82 & 78.57 & 63.55  \\
         & LR & 95.76  & 91.17 & \textbf{83.93} & 63.39  \\
         & SVM & 96.49 & 92.52 & \textbf{83.93} & 64.50\\
         &SLP & 95.16 & 91.58 & \textbf{83.93} & \textbf{66.30} \\
         & GDA & \textbf{96.16} & \textbf{92.79}  & 82.14 & 64.50 \\
    \bottomrule
    \end{tabular}
}
    \caption{ICL cannot be applied to tasks with a large number of classes (i.e., CLINC, Banking) or a lengthy inputs (i.e., CB, BoolQ). While our method inherits similar problem in PALP-D but we can apply PALP-T to linear probing methods. For each dataset, the best method is in bold. }
    \label{tab:largeclass}
\end{table}
\endgroup

\subsection{Full-data Results}
    Table \ref{tab:full-shot}, \ref{tab:largeclass} display the performance of different methodologies when the training data is fully available.
    Appending a simple template also displays a significant advantage even in a data-rich scenario, obtaining considerable improvements in accuracy regardless of the methods or the dataset.
    Notably, the performance of $k$-NN increases dramatically with the application of the template (PALP-T), which improved accuracy by 16\% on the Emotion dataset.

% --- figure start---
\begin{figure*}[t]
    \centering
    \begin{subfigure}{.23\linewidth}
        \centering
        \includegraphics[width=.99\linewidth]{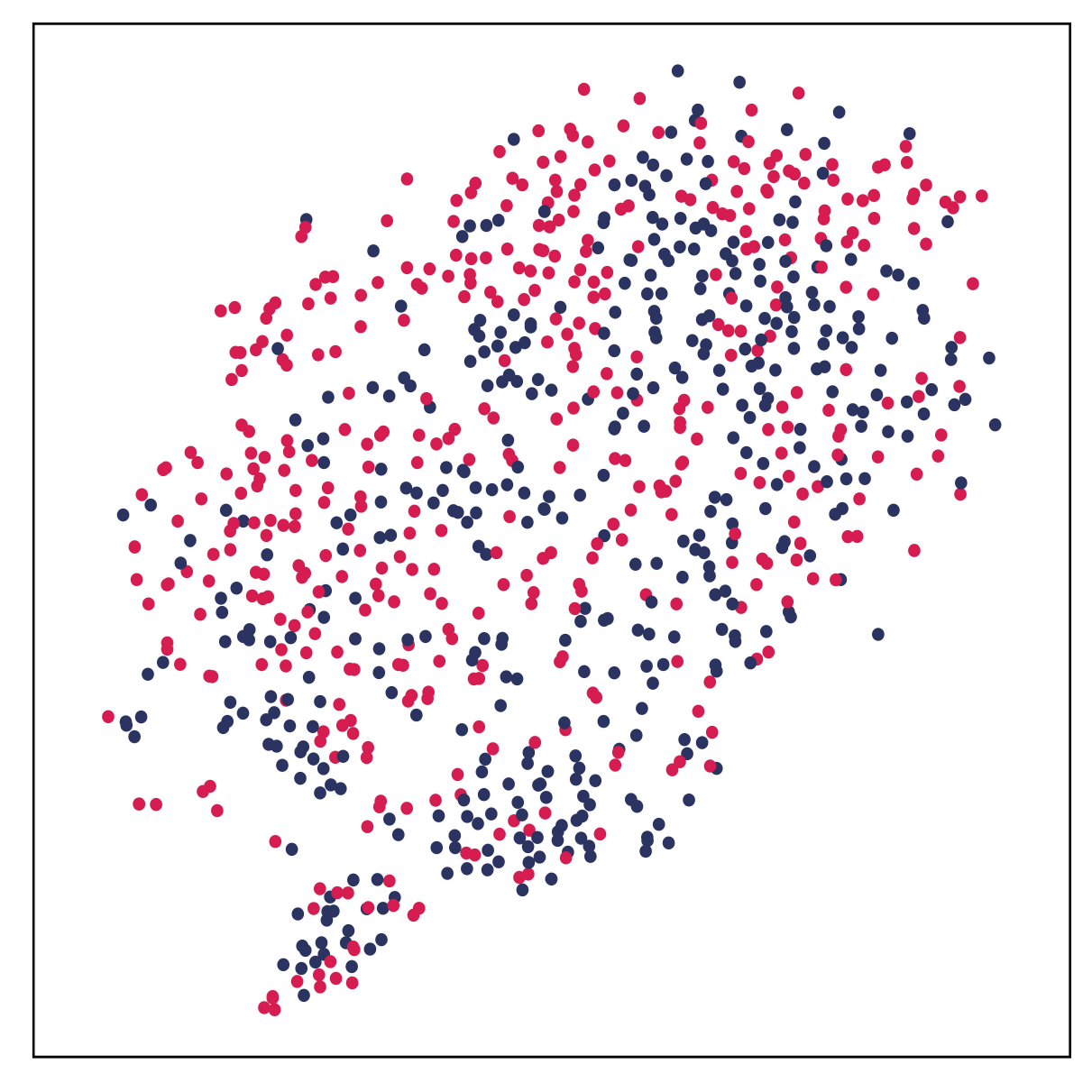}
        \caption{Baseline}
        \label{fig:tsne-vanilla}
    \end{subfigure}
    \begin{subfigure}{.23\linewidth}
        \centering
        \includegraphics[width=.99\linewidth]{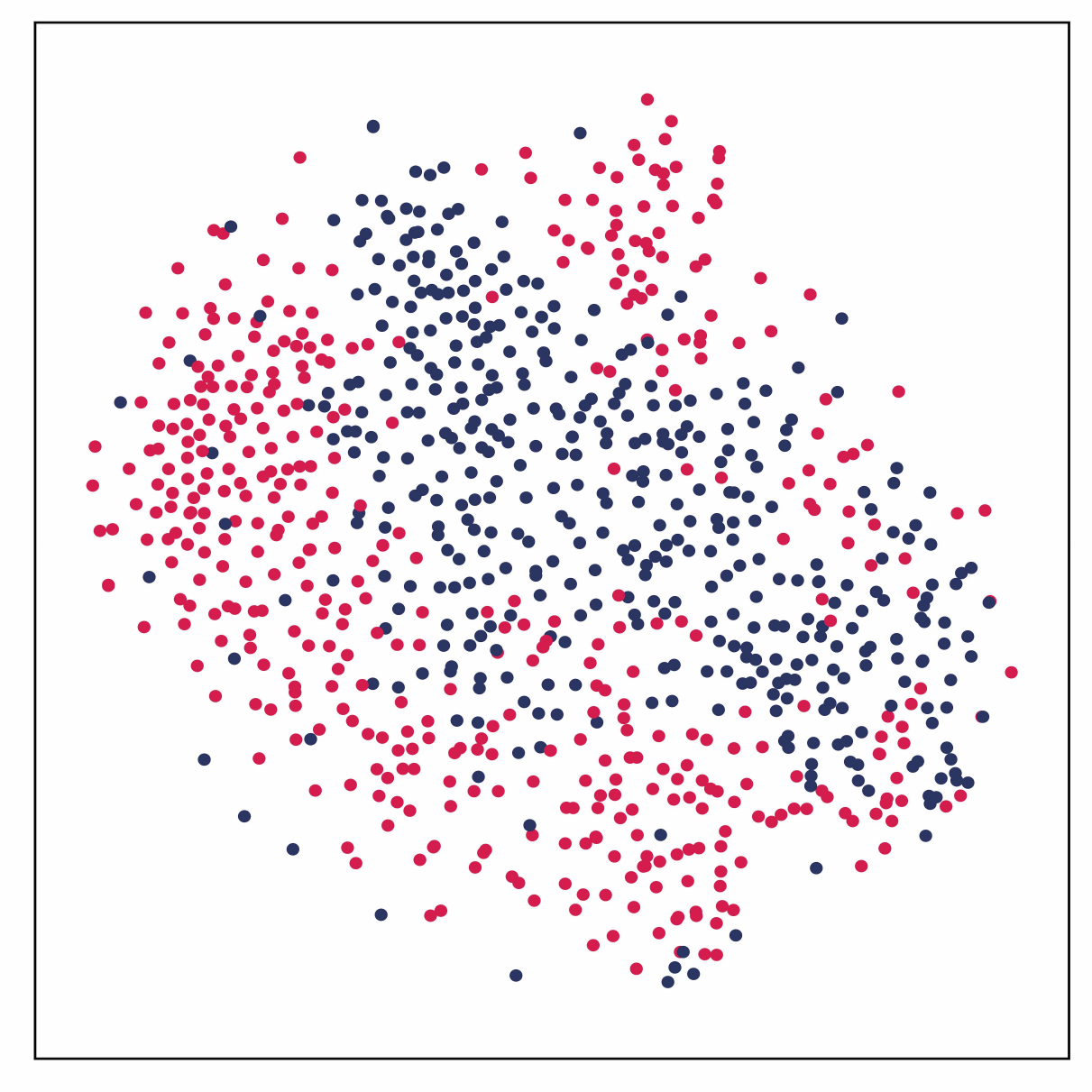}
        \caption{Template}
        \label{fig:tsne-template}
    \end{subfigure}
        \begin{subfigure}{.23\linewidth}
        \centering
        \includegraphics[width=.99\linewidth]{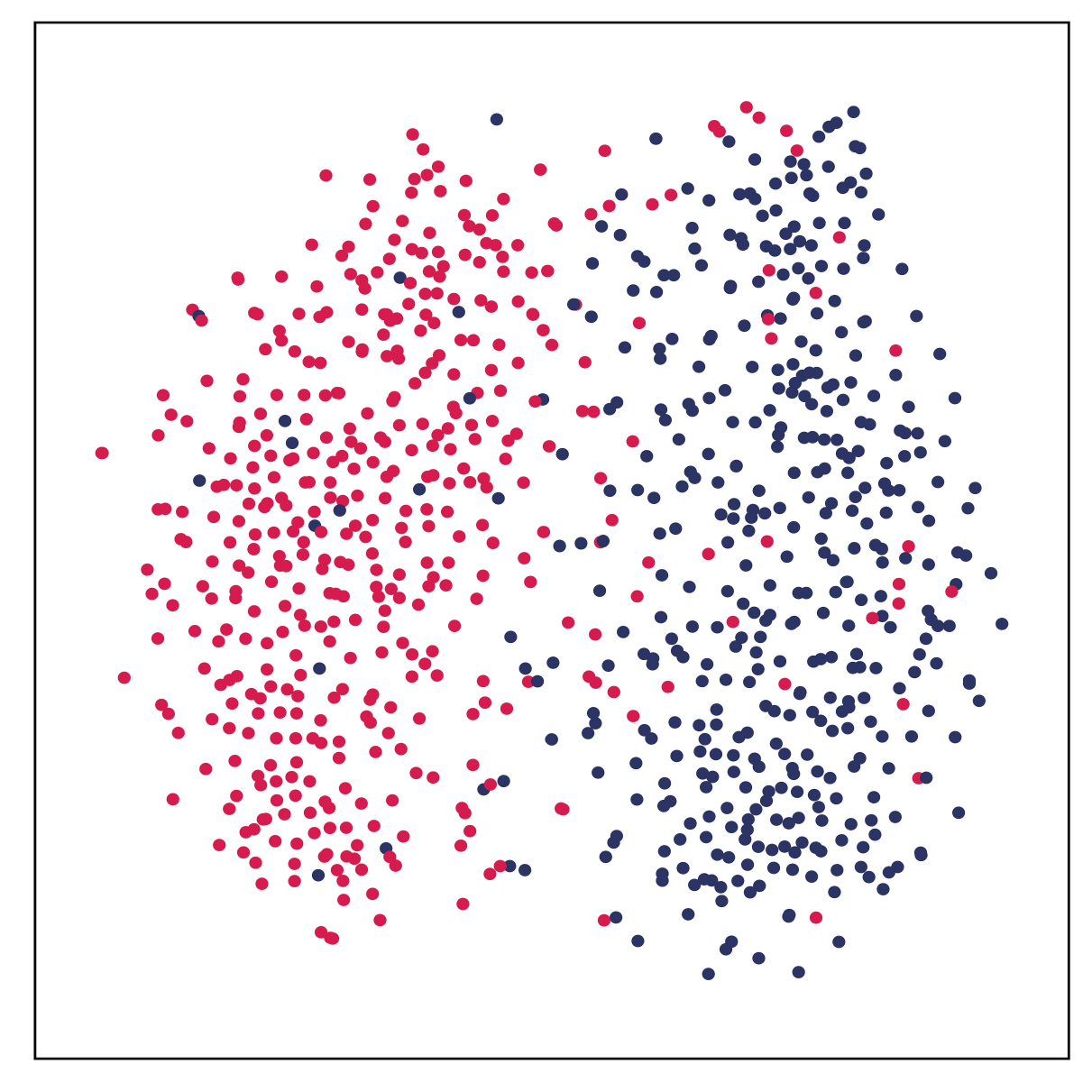}
        \caption{Demonstration-best}
        \label{fig:tsne-demonbest}
    \end{subfigure}
    \begin{subfigure}{.23\linewidth}
        \centering
        \includegraphics[width=.99\linewidth]{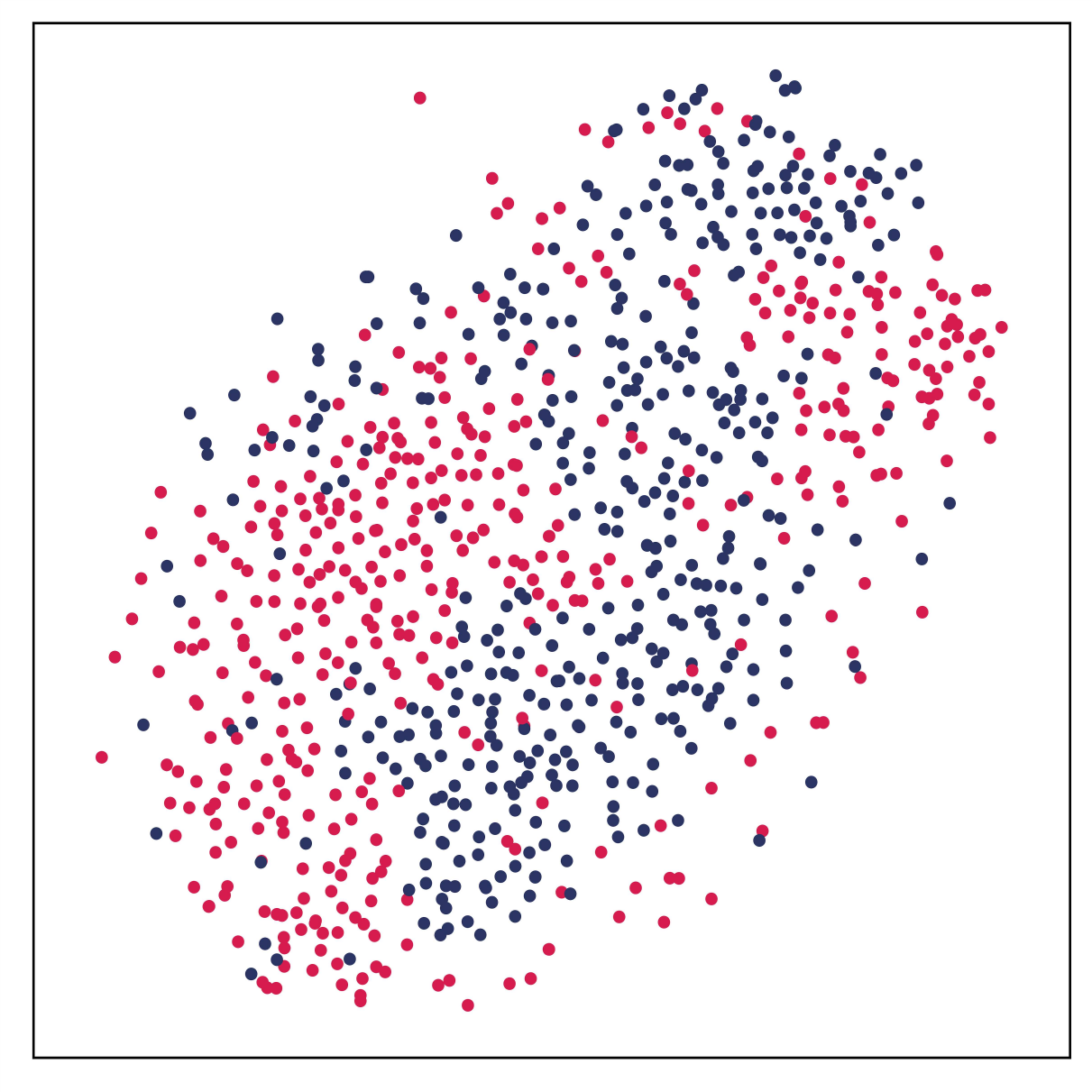}
        \caption{Demonstration-worst}
        \label{fig:tsne-demonworst}
    \end{subfigure}
    \par \medskip
    \caption{$t$-SNE visualization of SST2 representation from GPT-J. Adding understandable prompts to the input can reshape the representation into a more task-specially clustered form without any supervision. Demonstration-best is the representation obtained by attaching the demonstration that showed the best performance, and Demonstration-worst is the opposite. }
    \label{fig:t-sne}
\end{figure*}

\begin{table*}[t]

        \centering
        \begingroup
        \setlength{\tabcolsep}{6pt} % Default value: 6pt
        
        \begin{tabular}{c|l|l}
            \toprule
                Acc & \multicolumn{1}{c|}{Prefix ($\tau$)} & \multicolumn{1}{c}{Visualization} \\
            \midrule
                \begin{tabular}[c]{@{}c@{}}Max (84.86)\end{tabular} & \begin{tabular}[c]{@{}l@{}}Sentence 1: awful movie. \textbackslash \textbackslash Sentiment: negative \\Sentence 1: soulful , scathing and joyous. \textbackslash \textbackslash Sentiment: positive\end{tabular} & Fig. \ref{fig:tsne-demonbest} \\ \hline
                
                \begin{tabular}[c]{@{}c@{}}Min (54.62)\end{tabular} & \begin{tabular}[c]{@{}l@{}}Sentence 1: without any passion. \textbackslash \textbackslash Sentiment: negative \\Sentence 1: , incoherence and sub-sophomoric. \textbackslash \textbackslash Sentiment: positive\end{tabular} & Fig. \ref{fig:tsne-demonworst} \\
            \bottomrule
        \end{tabular}

       \endgroup
       \caption{Best and worst performing example prefixes from SST2.}
       \label{tab:week1}

     \label{tab:temps}
\end{table*}

    However, the method of appending demonstration (PALP-D) has more cons than pros in a data-abundant scenario.
    First, while  PALP-D often performs similarly to or better than PALP-T, they also sometimes yield worse scores than PALP-T, leading to a similar performance on average.
    Speaking otherwise, PALP-D only makes the input more lengthy and entails much higher inference costs compared to PALP-T in a data-abundant scenario.
    Moreover, PALP-D is infeasible to be applied to some tasks inheriting the limitations of ICL, as can be seen in Table \ref{tab:largeclass}: tasks with a large number of classes (i.e., CLINC, Banking), or tasks with long inputs (i.e., CB, BoolQ).

    Nevertheless, PALP greatly minimizes the performance gap between \textit{white-box} tuning methods, such as Adapter or full fine-tuning, and \textit{black-box} tuning methods, which is around 7\% with baseline linear probing methods, while our approach narrows this gap to nearly 4\%.
    In particular, our method outperforms or reaches statistically equivalent performance to \textit{white-box tuning} methods in some tasks, such as rotten tomatoes, TREC, and CLINC.

\section{Analysis \& Ablations}


\begingroup
\setlength{\tabcolsep}{15pt} % Default value: 6pt

\begin{table}[t]
\centering
\resizebox{1 \columnwidth}{!}{%
    \begin{tabular}{c|c|ccc}
    \toprule
        \multicolumn{5}{c}{GPT-2 (Large) 4-shot per class} \\
    \midrule
    \multicolumn{2}{c|}{Method} & SST2 & AG & MNLI \\
    \midrule
        \multirow{5}{*}{B} & k-NN & 50.34 & 26.59 & 33.86 \\
         & LR & 51.81 & 46.06 & 33.22 \\
         & SVM & 50.99 & 36.18 & 34.24 \\
         &SLP & 49.77 & 43.56 & 35.05 \\
         & GDA & 50.23 & 28.23 & 33.87 \\
    \midrule
        \multirow{5}{*}{T} & k-NN & 53.53 & 42.49 & 34.35 \\
         & LR & 53.23 & 52.36 & 32.8 \\
         & SVM & 51.77 & 44.10 & 34.04 \\
         &SLP & 52.27 & 43.23 & 34.97 \\
         & GDA & 54.91 & 35.51 & 33.26 \\
    \midrule
        \multirow{5}{*}{D} & k-NN & 56.54 & 63.44 & 35.93 \\
         & LR & 58.11 & 69.87 & 37.44 \\
         & SVM & 57.86 & \textbf{70.88} & \textbf{37.84} \\
         &SLP & 55.02 & 70.58 & 37.09 \\
         & GDA & \textbf{58.56} & 61.78 & 36.54 \\
     \bottomrule
    \end{tabular}
}
    \caption{Results on GPT-2 (Large) in 4-shot per class setting. B, T, and D refers to a baseline, template, and demonstration individually. Our method is transferable to smaller model.}
    \label{tab:gpt2}
\end{table}
\endgroup    

\subsection{Application to Small PLM}
    In this subsection, we examine our method for relatively small PLM to verify whether our approach is transferrable to small language models.
    Namely, we report the performance of 3 different tasks (sentiment analysis, natural language inference, and multiclass classification) on GPT-2 large \cite{radford2019language} in a 4-shot per class setting.
    Table \ref{tab:gpt2} summarizes the performance.
    Similar to previous experiments, our method mainly shows considerable performance gain on single sentence tasks and relatively small improvement on challenging tasks like sentence pair tasks.
    To summarize, PALP also benefits smaller language models, unlike ICL, which is known to yield poor performance or be unable to apply them to relatively small language models. 
    However, the performance improvement is less significant with smaller PLMs since our methodology depends solely on the capability of the language model.

\subsection{Dataset Visualization}

    In this experiment, we visualize the representation space when the differing input pre-processing method is applied to the input.
    Figure \ref{fig:t-sne} is the result of the $t$-SNE visualized representation of the SST2 task on GPT-J.
    Demonstration-best is the representation obtained by attaching the demonstration that showed the best performance, and Demonstration-worst is the opposite.
    Table \ref{tab:week1} summarizes actual examples of demonstrations and accuracy of the aforementioned best and worst cases.

    Consistent with the experimental results, we confirmed that PLM could extract more distinctive representations when appropriate templates or demonstrations are concatenated to the input of interest.
    The fact that a more meaningful representation can be drawn by applying a template is understandable and quite intuitive, as research has already shown that using a template can benefit fine-tuning performance \cite{liu2021gpt,schick2020exploiting, schick2020s}.
    What is even more intriguing is that adding demonstrations to the front can promote language models to derive a more task-specific and form a more distinguishable representation cluster.
    While the degree of improvement varies significantly, depending on the selected demonstrations, even concatenating the poorest performing demonstration yields a more clustered representation than the baseline result indicating the language model's capability to learn from the context of the input \cite{min2022rethinking}.
    Although we did not specifically identify which demonstrations were more helpful and which were not, we found that mislabeled demonstrations can hurt the overall quality of the representations.
    As can be seen in Table \ref{tab:week1}, the second demonstration of the worst demonstrations is wrongly labeled, where we conjecture is the cause of the poor performance as advocated in a recent study \citet{kim2022ground}  that the mapping of input sentence and ground-truth label space can be crucial.

    \section{Related Work}
        Large language models such as GPT-3 \cite{brown2020language} and ERNIE 3.0 \cite{sun2021ernie} are often released as black-box APIs due to commercial considerations and the potential risk of misuse. 
        Thus, users are unable to train those pre-trained models with the traditional transferring paradigm (i.e., fine-tuning). 
        Even in some cases where the weights of the pre-trained model are accessible \cite{zhang2022opt, scao2022bloom}, it may not be possible for many researchers to fine-tune those models due to the enormous resource they require. 
        Several studies were proposed to circumvent the problems as mentioned earlier:

    \subsection{Black-box Tuning}
        Black-box tuning is a methodology that makes use of the target model without internal model access (e.g., gradient, middle layer representations, or weights).
        As such, black-box tuning methods usually train a lightweight discriminator on top of pre-trained models or optimize input prompt in a derivative-free fashion since the gradient of the pre-trained language models is unavailable. 
        Specifically, \citet{diao2022black, sun2022black} attempts to find optimal prompt input without utilizes the natural evolution strategy (NES) to find better prompts for black-box models instead of using natural NES to fool the model as in black-box adversarial attacks.
        \citet{sun2022black} adopted a covariance matrix adaptation evolution strategy to perform optimization in a randomly generated small subspace, which is effective due to the low intrinsic dimensionality of large language models \cite{aghajanyan2020intrinsic, qin2021exploring}.
    
    \subsection{In-Context Learning}
        ICL \cite{brown2020language} is an alternative to the gradient-based tuning method. 
        It is a novel transferring method that derives answers via conditioning appropriate prompts or often concatenating the training data. 
        ICL is drawing explosive interest in the field of NLP due to their strong generalizability among many different tasks, from traditional natural language understanding tasks, including sentiment analysis and natural language inference \cite{wang2018glue}, to extreme ones, such as code generations \cite{poesia2022synchromesh} or mathematical problems \cite{henighan2020scaling}.
        
         As the underlying mechanism of ICL astonished the NLP field and has reminisced the capability of PLMs, a plethora of works has been proposed to utilize and understand ICL better. Studies include advanced ICL methods maximizing the downstream performance \cite{zhao2021calibrate, min2021noisy, holtzman2021surface}, advanced methods of choosing example data \cite{liu2021makes, lu2021fantastically, rubin2021learning}, understanding the limitation of ICL \cite{liu2021makes, lu2021fantastically}, and understanding the underlying mechanism of ICL \cite{xie2021explanation, reynolds2021prompt, min2022rethinking, razeghi2022impact, kim2022ground}.

\section{Conclusion }
    In this paper, we showed that providing task descriptions or demonstrations can enforce PLM to yield more robust representations without additional adaptation of the model weights, allowing them to be used for lightweight linear probing as an alternative to in-context learning.
    In light of this finding, we proposed prompt-augmented linear probing, where we augmented data representations with ICL-style crafted inputs. 
    Our integrated approach is scalable with the available training data and the size of the language model.
    PALP obtains comparable results to ICL in the data-hungry scenario and comparable results to fine-tuning in the data-abundant scenario with little training overhead, potentially making PALP a strong alternative in various situations.

    In our follow-up study, we will analyze how the additional prompt tokens (e.g., demonstrations or templates) affect the representation quality of the encapsulating input text.
    We are also interested in the effect of adopting self-supervised learning objectives, such as contrastive learning \cite{gao2021simcse},  to the shallow layers on top of the language model backbone, which might improve our method further.
    \section{Acknowledgements}

    This work was mainly supported by SNU-NAVER Hyperscale AI Center and partially by the Institute of Information \& communications Technology Planning \& Evaluation (IITP) grant funded by the Korean government (MSIT) No.2020-0-01373, Artificial Intelligence Graduate School Program (Hanyang University), and No.2021-0-01343, Artificial Intelligence Graduate School Program (Seoul National University)].
    Lastly, we would like to express gratitude to Kyunghyun Cho and the anonymous reviewers for their precious feedback.

\bibliography{aaai23}

\begin{thebibliography}{38}
\providecommand{\natexlab}[1]{#1}

\bibitem[{Aghajanyan, Gupta, and Zettlemoyer(2021)}]{aghajanyan2020intrinsic}
Aghajanyan, A.; Gupta, S.; and Zettlemoyer, L. 2021.
\newblock Intrinsic Dimensionality Explains the Effectiveness of Language Model
  Fine-Tuning.
\newblock In Zong, C.; Xia, F.; Li, W.; and Navigli, R., eds.,
  \emph{Proceedings of the 59th Annual Meeting of the Association for
  Computational Linguistics, {ACL}}.

\bibitem[{Ben-David, Oved, and Reichart(2022)}]{ben2021pada}
Ben-David, E.; Oved, N.; and Reichart, R. 2022.
\newblock {PADA}: Example-based Prompt Learning for on-the-fly Adaptation to
  Unseen Domains.
\newblock \emph{Transactions of the Association for Computational Linguistics}.

\bibitem[{Brown et~al.(2020)Brown, Mann, Ryder, Subbiah, Kaplan, Dhariwal,
  Neelakantan, Shyam, Sastry, Askell et~al.}]{brown2020language}
Brown, T.; Mann, B.; Ryder, N.; Subbiah, M.; Kaplan, J.~D.; Dhariwal, P.;
  Neelakantan, A.; Shyam, P.; Sastry, G.; Askell, A.; et~al. 2020.
\newblock Language models are few-shot learners.
\newblock \emph{Advances in neural information processing systems}, 33:
  1877--1901.

\bibitem[{Chowdhery et~al.(2022)Chowdhery, Narang, Devlin, Bosma, Mishra,
  Roberts, Barham, Chung, Sutton, Gehrmann et~al.}]{chowdhery2022palm}
Chowdhery, A.; Narang, S.; Devlin, J.; Bosma, M.; Mishra, G.; Roberts, A.;
  Barham, P.; Chung, H.~W.; Sutton, C.; Gehrmann, S.; et~al. 2022.
\newblock Palm: Scaling language modeling with pathways.
\newblock \emph{arXiv preprint arXiv:2204.02311}.

\bibitem[{Diao et~al.(2022)Diao, Li, Lin, Huang, and Zhang}]{diao2022black}
Diao, S.; Li, X.; Lin, Y.; Huang, Z.; and Zhang, T. 2022.
\newblock Black-box prompt learning for pre-trained language models.
\newblock \emph{arXiv preprint arXiv:2201.08531}.

\bibitem[{Fedus, Zoph, and Shazeer(2022)}]{fedus2021switch}
Fedus, W.; Zoph, B.; and Shazeer, N. 2022.
\newblock Switch Transformers: Scaling to Trillion Parameter Models with Simple
  and Efficient Sparsity.
\newblock \emph{Journal of Machine Learning Research}, 23(120): 1--39.

\bibitem[{Gao, Yao, and Chen(2021)}]{gao2021simcse}
Gao, T.; Yao, X.; and Chen, D. 2021.
\newblock SimCSE: Simple Contrastive Learning of Sentence Embeddings.
\newblock In \emph{Proceedings of the 2021 Conference on Empirical Methods in
  Natural Language Processing}.

\bibitem[{Henighan et~al.(2020)Henighan, Kaplan, Katz, Chen, Hesse, Jackson,
  Jun, Brown, Dhariwal, Gray et~al.}]{henighan2020scaling}
Henighan, T.; Kaplan, J.; Katz, M.; Chen, M.; Hesse, C.; Jackson, J.; Jun, H.;
  Brown, T.~B.; Dhariwal, P.; Gray, S.; et~al. 2020.
\newblock Scaling laws for autoregressive generative modeling.
\newblock \emph{arXiv preprint arXiv:2010.14701}.

\bibitem[{Hoffmann et~al.(2022)Hoffmann, Borgeaud, Mensch, Buchatskaya, Cai,
  Rutherford, Casas, Hendricks, Welbl, Clark et~al.}]{hoffmann2022training}
Hoffmann, J.; Borgeaud, S.; Mensch, A.; Buchatskaya, E.; Cai, T.; Rutherford,
  E.; Casas, D. d.~L.; Hendricks, L.~A.; Welbl, J.; Clark, A.; et~al. 2022.
\newblock Training Compute-Optimal Large Language Models.
\newblock \emph{arXiv preprint arXiv:2203.15556}.

\bibitem[{Holtzman et~al.(2021)Holtzman, West, Shwartz, Choi, and
  Zettlemoyer}]{holtzman2021surface}
Holtzman, A.; West, P.; Shwartz, V.; Choi, Y.; and Zettlemoyer, L. 2021.
\newblock Surface Form Competition: Why the Highest Probability Answer Isn't
  Always Right.
\newblock In \emph{Proceedings of the 2021 Conference on Empirical Methods in
  Natural Language Processing, {EMNLP}}.

\bibitem[{Houlsby et~al.(2019)Houlsby, Giurgiu, Jastrzebski, Morrone,
  De~Laroussilhe, Gesmundo, Attariyan, and Gelly}]{houlsby2019parameter}
Houlsby, N.; Giurgiu, A.; Jastrzebski, S.; Morrone, B.; De~Laroussilhe, Q.;
  Gesmundo, A.; Attariyan, M.; and Gelly, S. 2019.
\newblock Parameter-efficient transfer learning for NLP.
\newblock In \emph{International Conference on Machine Learning}.

\bibitem[{Kaplan et~al.(2020)Kaplan, McCandlish, Henighan, Brown, Chess, Child,
  Gray, Radford, Wu, and Amodei}]{kaplan2020scaling}
Kaplan, J.; McCandlish, S.; Henighan, T.; Brown, T.~B.; Chess, B.; Child, R.;
  Gray, S.; Radford, A.; Wu, J.; and Amodei, D. 2020.
\newblock Scaling Laws for Neural Language Models.
\newblock \emph{CoRR}.

\bibitem[{Liu et~al.(2022{\natexlab{a}})Liu, Shen, Zhang, Dolan, Carin, and
  Chen}]{liu2021makes}
Liu, J.; Shen, D.; Zhang, Y.; Dolan, B.; Carin, L.; and Chen, W.
  2022{\natexlab{a}}.
\newblock What Makes Good In-Context Examples for GPT-3?
\newblock In \emph{Proceedings of Deep Learning Inside Out: The 3rd Workshop on
  Knowledge Extraction and Integration for Deep Learning Architectures,
  DeeLIO@ACL}.

\bibitem[{Liu et~al.(2022{\natexlab{b}})Liu, Yuan, Fu, Jiang, Hayashi, and
  Neubig}]{liu2021pre}
Liu, P.; Yuan, W.; Fu, J.; Jiang, Z.; Hayashi, H.; and Neubig, G.
  2022{\natexlab{b}}.
\newblock Pre-Train, Prompt, and Predict: A Systematic Survey of Prompting
  Methods in Natural Language Processing.
\newblock \emph{ACM Comput. Surv.}

\bibitem[{Liu et~al.(2021)Liu, Zheng, Du, Ding, Qian, Yang, and
  Tang}]{liu2021gpt}
Liu, X.; Zheng, Y.; Du, Z.; Ding, M.; Qian, Y.; Yang, Z.; and Tang, J. 2021.
\newblock GPT Understands, Too.
\newblock \emph{arXiv preprint arXiv:2103.10385}.

\bibitem[{Lu et~al.(2022)Lu, Bartolo, Moore, Riedel, and
  Stenetorp}]{lu2021fantastically}
Lu, Y.; Bartolo, M.; Moore, A.; Riedel, S.; and Stenetorp, P. 2022.
\newblock Fantastically Ordered Prompts and Where to Find Them: Overcoming
  Few-Shot Prompt Order Sensitivity.
\newblock In \emph{Proceedings of the 60th Annual Meeting of the Association
  for Computational Linguistics, {ACL}}.

\bibitem[{Min et~al.(2022{\natexlab{a}})Min, Lewis, Hajishirzi, and
  Zettlemoyer}]{min2021noisy}
Min, S.; Lewis, M.; Hajishirzi, H.; and Zettlemoyer, L. 2022{\natexlab{a}}.
\newblock Noisy Channel Language Model Prompting for Few-Shot Text
  Classification.
\newblock In \emph{Proceedings of the 60th Annual Meeting of the Association
  for Computational Linguistics, {ACL}}.

\bibitem[{Min et~al.(2022{\natexlab{b}})Min, Lyu, Holtzman, Artetxe, Lewis,
  Hajishirzi, and Zettlemoyer}]{min2022rethinking}
Min, S.; Lyu, X.; Holtzman, A.; Artetxe, M.; Lewis, M.; Hajishirzi, H.; and
  Zettlemoyer, L. 2022{\natexlab{b}}.
\newblock Rethinking the Role of Demonstrations: What Makes In-Context Learning
  Work?
\newblock \emph{Proceedings of the 2020 Conference on Empirical Methods in
  Natural Language Processing (EMNLP)}.

\bibitem[{Petroni et~al.(2019)Petroni, Rockt{\"{a}}schel, Riedel, Lewis,
  Bakhtin, Wu, and Miller}]{petroni2019language}
Petroni, F.; Rockt{\"{a}}schel, T.; Riedel, S.; Lewis, P. S.~H.; Bakhtin, A.;
  Wu, Y.; and Miller, A.~H. 2019.
\newblock Language Models as Knowledge Bases?
\newblock In Inui, K.; Jiang, J.; Ng, V.; and Wan, X., eds., \emph{Proceedings
  of the 2019 Conference on Empirical Methods in Natural Language Processing,
  {EMNLP}}.

\bibitem[{Poesia et~al.(2022)Poesia, Polozov, Le, Tiwari, Soares, Meek, and
  Gulwani}]{poesia2022synchromesh}
Poesia, G.; Polozov, A.; Le, V.; Tiwari, A.; Soares, G.; Meek, C.; and Gulwani,
  S. 2022.
\newblock Synchromesh: Reliable Code Generation from Pre-trained Language
  Models.
\newblock In \emph{The Tenth International Conference on Learning
  Representations, {ICLR}}.

\bibitem[{Qin et~al.(2021)Qin, Wang, Su, Lin, Ding, Liu, Li, Hou, Li, Sun
  et~al.}]{qin2021exploring}
Qin, Y.; Wang, X.; Su, Y.; Lin, Y.; Ding, N.; Liu, Z.; Li, J.; Hou, L.; Li, P.;
  Sun, M.; et~al. 2021.
\newblock Exploring low-dimensional intrinsic task subspace via prompt tuning.
\newblock \emph{arXiv preprint arXiv:2110.07867}.

\bibitem[{Radford et~al.(2019)Radford, Wu, Child, Luan, Amodei, Sutskever
  et~al.}]{radford2019language}
Radford, A.; Wu, J.; Child, R.; Luan, D.; Amodei, D.; Sutskever, I.; et~al.
  2019.
\newblock Language models are unsupervised multitask learners.
\newblock \emph{OpenAI blog}, 1(8): 9.

\bibitem[{Razeghi et~al.(2022)Razeghi, Logan~IV, Gardner, and
  Singh}]{razeghi2022impact}
Razeghi, Y.; Logan~IV, R.~L.; Gardner, M.; and Singh, S. 2022.
\newblock Impact of pretraining term frequencies on few-shot reasoning.
\newblock \emph{arXiv preprint arXiv:2202.07206}.

\bibitem[{Reynolds and McDonell(2021)}]{reynolds2021prompt}
Reynolds, L.; and McDonell, K. 2021.
\newblock Prompt programming for large language models: Beyond the few-shot
  paradigm.
\newblock In \emph{Extended Abstracts of the 2021 CHI Conference on Human
  Factors in Computing Systems}, 1--7.

\bibitem[{Rubin, Herzig, and Berant(2022)}]{rubin2021learning}
Rubin, O.; Herzig, J.; and Berant, J. 2022.
\newblock Learning To Retrieve Prompts for In-Context Learning.
\newblock In \emph{Proceedings of the 2022 Conference of the North American
  Chapter of the Association for Computational Linguistics, {NAACL}}.

\bibitem[{Scao et~al.(2022)Scao, Fan, Akiki, Pavlick, Ili{\'c}, Hesslow,
  Castagn{\'e}, Luccioni, Yvon, Gall{\'e} et~al.}]{scao2022bloom}
Scao, T.~L.; Fan, A.; Akiki, C.; Pavlick, E.; Ili{\'c}, S.; Hesslow, D.;
  Castagn{\'e}, R.; Luccioni, A.~S.; Yvon, F.; Gall{\'e}, M.; et~al. 2022.
\newblock BLOOM: A 176B-Parameter Open-Access Multilingual Language Model.
\newblock \emph{arXiv preprint arXiv:2211.05100}.

\bibitem[{Schick and Sch{\"{u}}tze(2021{\natexlab{a}})}]{schick2020exploiting}
Schick, T.; and Sch{\"{u}}tze, H. 2021{\natexlab{a}}.
\newblock Exploiting Cloze-Questions for Few-Shot Text Classification and
  Natural Language Inference.
\newblock In Merlo, P.; Tiedemann, J.; and Tsarfaty, R., eds.,
  \emph{Proceedings of the 16th Conference of the European Chapter of the
  Association for Computational Linguistics, {EACL}}.

\bibitem[{Schick and Sch{\"{u}}tze(2021{\natexlab{b}})}]{schick2020s}
Schick, T.; and Sch{\"{u}}tze, H. 2021{\natexlab{b}}.
\newblock It's Not Just Size That Matters: Small Language Models Are Also
  Few-Shot Learners.
\newblock In \emph{Proceedings of the 2021 Conference of the North American
  Chapter of the Association for Computational Linguistics, {NAACL}}.

\bibitem[{Shwartz et~al.(2020)Shwartz, West, Le~Bras, Bhagavatula, and
  Choi}]{shwartz2020unsupervised}
Shwartz, V.; West, P.; Le~Bras, R.; Bhagavatula, C.; and Choi, Y. 2020.
\newblock Unsupervised Commonsense Question Answering with Self-Talk.
\newblock In \emph{Proceedings of the 2020 Conference on Empirical Methods in
  Natural Language Processing (EMNLP)}, 4615--4629.

\bibitem[{Sun et~al.(2022)Sun, Shao, Qian, Huang, and Qiu}]{sun2022black}
Sun, T.; Shao, Y.; Qian, H.; Huang, X.; and Qiu, X. 2022.
\newblock Black-Box Tuning for Language-Model-as-a-Service.
\newblock In Chaudhuri, K.; Jegelka, S.; Song, L.; Szepesv{\'{a}}ri, C.; Niu,
  G.; and Sabato, S., eds., \emph{International Conference on Machine Learning,
  {ICML}}.

\bibitem[{Sun et~al.(2021)Sun, Wang, Feng, Ding, Pang, Shang, Liu, Chen, Zhao,
  Lu et~al.}]{sun2021ernie}
Sun, Y.; Wang, S.; Feng, S.; Ding, S.; Pang, C.; Shang, J.; Liu, J.; Chen, X.;
  Zhao, Y.; Lu, Y.; et~al. 2021.
\newblock Ernie 3.0: Large-scale knowledge enhanced pre-training for language
  understanding and generation.
\newblock \emph{arXiv preprint arXiv:2107.02137}.

\bibitem[{Vaswani et~al.(2017)Vaswani, Shazeer, Parmar, Uszkoreit, Jones,
  Gomez, Kaiser, and Polosukhin}]{vaswani2017attention}
Vaswani, A.; Shazeer, N.; Parmar, N.; Uszkoreit, J.; Jones, L.; Gomez, A.~N.;
  Kaiser, {\L}.; and Polosukhin, I. 2017.
\newblock Attention is all you need.
\newblock In \emph{Advances in neural information processing systems},
  5998--6008.

\bibitem[{Wang et~al.(2019)Wang, Singh, Michael, Hill, Levy, and
  Bowman}]{wang2018glue}
Wang, A.; Singh, A.; Michael, J.; Hill, F.; Levy, O.; and Bowman, S.~R. 2019.
\newblock {GLUE:} {A} Multi-Task Benchmark and Analysis Platform for Natural
  Language Understanding.
\newblock In \emph{7th International Conference on Learning Representations,
  {ICLR}}.

\bibitem[{Wang and Komatsuzaki(2021)}]{gpt-j}
Wang, B.; and Komatsuzaki, A. 2021.
\newblock {GPT-J-6B: A 6 Billion Parameter Autoregressive Language Model}.
\newblock \url{https://github.com/kingoflolz/mesh-transformer-jax}.
\newblock Accessed: 2021-05.

\bibitem[{Xie et~al.(2022)Xie, Raghunathan, Liang, and Ma}]{xie2021explanation}
Xie, S.~M.; Raghunathan, A.; Liang, P.; and Ma, T. 2022.
\newblock An Explanation of In-context Learning as Implicit Bayesian Inference.
\newblock In \emph{The Tenth International Conference on Learning
  Representations, {ICLR}}.

\bibitem[{Yoo et~al.(2022)Yoo, Kim, Kim, Cho, Jo, Lee, Lee, and
  Kim}]{kim2022ground}
Yoo, K.~M.; Kim, J.; Kim, H.~J.; Cho, H.; Jo, H.; Lee, S.-W.; Lee, S.-g.; and
  Kim, T. 2022.
\newblock Ground-Truth Labels Matter: A Deeper Look into Input-Label
  Demonstrations.
\newblock \emph{Proceedings of the 2020 Conference on Empirical Methods in
  Natural Language Processing (EMNLP)}.

\bibitem[{Zhang et~al.(2022)Zhang, Roller, Goyal, Artetxe, Chen, Chen, Dewan,
  Diab, Li, Lin et~al.}]{zhang2022opt}
Zhang, S.; Roller, S.; Goyal, N.; Artetxe, M.; Chen, M.; Chen, S.; Dewan, C.;
  Diab, M.; Li, X.; Lin, X.~V.; et~al. 2022.
\newblock Opt: Open pre-trained transformer language models.
\newblock \emph{arXiv preprint arXiv:2205.01068}.

\bibitem[{Zhao et~al.(2021)Zhao, Wallace, Feng, Klein, and
  Singh}]{zhao2021calibrate}
Zhao, Z.; Wallace, E.; Feng, S.; Klein, D.; and Singh, S. 2021.
\newblock Calibrate Before Use: Improving Few-shot Performance of Language
  Models.
\newblock In Meila, M.; and Zhang, T., eds., \emph{Proceedings of the 38th
  International Conference on Machine Learning, {ICML}}.

\end{thebibliography}

\end{document}